\documentclass{article}

\usepackage[final]{neurips_2022}

\usepackage[utf8]{inputenc} % allow utf-8 input
\usepackage[T1]{fontenc}    % use 8-bit T1 fonts
\usepackage{hyperref}       % hyperlinks
\usepackage{url}            % simple URL typesetting
\usepackage{booktabs}       % professional-quality tables
\usepackage{amsfonts}       % blackboard math symbols
\usepackage{nicefrac}       % compact symbols for 1/2, etc.
\usepackage{microtype}      % microtypography
\usepackage{xcolor}

\usepackage{graphicx}
\usepackage{amsmath}
\usepackage{amsthm}
\usepackage{booktabs}
\usepackage{amsfonts}
\usepackage{subcaption}
\usepackage{amssymb}
\usepackage{multirow}
\newcommand{\cNP}{{NP}}
\newcommand{\cP}{{P}}
\usepackage{wrapfig}
\newtheorem{theorem}{Theorem}[section]

\newtheorem{corollary}[]{Corollary}
\newtheorem{definition}[]{Definition}

\title{The (Un)Scalability of Heuristic Approximators \\ for NP-Hard Search Problems}
\author{%
  Sumedh Pendurkar \\
  Department of Computer Science \& Engineering\\
  Texas A\&M University\\
  \texttt{sumedhpendurkar@tamu.edu} 
  \And
  Taoan Huang \\
  Department of Computer Science\\
  University of Southern California\\
  \texttt{taoanhua@usc.edu}
  \And
  Sven Koenig \\
  Department of Computer Science\\
  University of Southern California\\
  \texttt{skoenig@usc.edu}
  \And
  Guni Sharon \\
  Department of Computer Science \& Engineering\\
  Texas A\&M University\\
  \texttt{guni@tamu.edu} 
}

\begin{document}

\maketitle

\begin{abstract}

The A* algorithm is commonly used to solve \cNP-hard combinatorial optimization problems. When provided with a completely informed heuristic function, A* solves many \cNP-hard minimum-cost path problems in time polynomial in the branching factor and the number of edges in a minimum-cost path. Thus, approximating their completely informed heuristic functions with high precision is \cNP-hard. We therefore examine recent publications that propose the use of neural networks for this purpose. We support our claim that these approaches do not scale to large instance sizes both theoretically and experimentally. Our first experimental results for three representative \cNP-hard minimum-cost path problems suggest that using neural networks to approximate completely informed heuristic functions with high precision might result in network sizes that scale exponentially in the instance sizes. The research community might thus benefit from investigating other ways of integrating heuristic search with machine learning.

\end{abstract}

\section{Introduction}

Solving combinatorial optimization problems is often \cNP-hard \citep{papadimitriou1998combinatorial}, in which case current algorithms are not able to solve them in time polynomial in their instance sizes. If the problems are \cNP-complete, then it is unknown whether they can be solved in polynomial time \citep{cook2006p}, which is one of the biggest open issues in computer science currently. Since algorithms that run in polynomial time might be unattainable for them, many researchers work on reducing the exponential runtime of current algorithms \citep{pearl1984heuristics}. A* uses heuristic functions for this purpose \citep{hart1968formal}. Informed heuristic functions were shown to reduce the runtime of A* by orders of magnitude \citep{goldenberg2014enhanced,felner2018adding}. In fact, completely informed heuristic functions make its runtime polynomial in the branching factor and the number of edges in a minimum-cost path and thus enable it to solve large instances of combinatorial optimization problems fast.

We expand on our initial discussion \citep{pendurkar2022discussion} by showing that machine learning can indeed be used to approximate completely informed heuristic functions with bounded function values to arbitrary precision for minimum-cost path problems with finite numbers of states. This insight supports recent publications that propose universal function approximators for this purpose \citep{mcaleer2018solving,agostinelli2019deepcubea,agostinelli2021search}. Unfortunately, we also show that approximating completely informed heuristic functions with high precision is NP-hard and thus might not scale to large instance sizes. Our first experimental results for three representative \cNP-hard minimum-cost path problems with different neural network topologies, loss thresholds, and loss functions suggest that using neural networks to approximate completely informed heuristic functions with high precision might result in network sizes that scale exponentially in the instance sizes.\footnote{See  \url{https://github.com/Pi-Star-Lab/unscalable-heuristic-approximator} for our code.}

Our conclusions regarding the scalability of finding approximations of completely informed heuristic functions can help direct the search community towards more promising avenues of integrating heuristic search algorithms with machine learning approaches when targeting \cNP-hard optimization problems.
% WE NEED TO CHECK CAREFULLY IN THE WHOLE PAPER WHERE TO SAY NP-HARD AND WHERE TO SAY NP-COMPLETE.

\section{Preliminaries}

\cP\ is the class of problems that are solvable in polynomial time with deterministic Turing machines. \cNP\ is the class of problems whose solutions can be verified in polynomial time. Thus, these problems are solvable in polynomial time with non-deterministic Turing machines. \cNP-hard is the class of problems to which every problem in \cNP\ can be reduced in polynomial time. Finally, \cNP-complete is the class of problems that are both \cNP-hard and in \cNP. Thus, \cP=\cNP\ if any of these problems is also in \cP. 

Many \cNP-hard problems can be reduced to the minimum-cost path problem on appropriate graphs $G=(S,E)$ with given start and goal states in $S$ \citep{gupta1992complexity,cormen2009introduction,bulteau2015pancake}, resulting in \cNP-hard minimum-cost path problems. Each vertex $s \in S$ represents a state. Each edge $e \in E$ represents an operator and is labeled with a cost $c(e) > 0$. The minimum-cost path from $s$ to the goal is $c^*(s):=\arg\min_{E' \subseteq E} \sum_{e \in E'} c(e)$, where $E'$ is an ordered set of edges leading from the start state to the goal state.     

A* is typically used to find a minimum-cost path from the start state to the goal state without constructing the full graph. The heuristic function $h: S \mapsto \mathbb{R}^+_0$ of A* estimates the goal distance of any state (that is, the cost of a minimum-cost path from the state to the goal state). For any given (consistent) heuristic function, A* expands (up to tie-breaking) the minimum number of states required for finding a minimum-cost path and proving its optimality~\citep{pearl1984heuristics}.

The completely informed heuristic function $h^*: S \mapsto \mathbb{R}^+_0$ returns $c^*(s)$ for any state $s$. For the completely informed heuristic function, A* (with appropriate tie-breaking) expands only the states along one minimum-cost path from the start state to the goal state, while also generating the successors of the expanded states. Its runtime is then polynomial in the branching factor (that is, the maximum number of successors of any state) and the number of edges in a minimum-cost path. Informed heuristic functions (that approximate the completely informed heuristic functions) have reduced the runtime of A* by orders of magnitude for \cNP-hard minimum-cost path problems  \citep{helmert2008accuracy}. Recent publications have suggested to use universal function approximators in form of deep neural networks~\citep{higgins2021generalizing} to approximate the completely informed heuristic function with high precision.

\section{Related Work}

Early on, \citet{arfaee2010bootstrap} used an iterative "Bootstrap Learning Heuristic" (BLH) approach to approximate the completely informed heuristic function. It repeats the following process: It runs A* with an approximation of the completely informed heuristic function (initially a weak one), obtains more informed estimates of the heuristic function values for the states on the path found by the search, and uses neural network-based supervised learning with these estimates to improve the approximation of the completely informed heuristic function. Later, \citet{mcaleer2018solving} applied deep (neural network-based) reinforcement learning instead of this iterative approach to improve a weak approximation of the completely informed heuristic function for the Rubik's cube over time. \citet{agostinelli2019deepcubea} improved this approach with their DeepCubeA algorithm, establishing the current state of the art. See Appendix~\ref{sec:related-summary} for a summary/comparison of existing approaches. 

\citet{orseau2018single} learned an approximation of the optimal policy instead of the completely informed heuristic function, where a policy maps each state to the operator that should be executed in it. Later, \citet{orseau2021policy} used a ``Policy-guided Heuristic Search" (PHS) approach that improves the performance of their previous approach by learning approximations of both the optimal policy and the completely informed heuristic function.

One can train universal function approximators that approximate completely informed heuristic functions with high precision. This includes feed-forward neural networks with suitable activation functions and one hidden layer if there are sufficiently many neurons in the hidden layer \citep{cybenko1989approximation,hornik1989multilayer,pinkus1999approximation}, we denote these as ``fixed-depth" neural networks. 
It also includes feed-forward neural networks with suitable activation functions and a fixed number of neurons per hidden layer if there are sufficiently many hidden layers and the number of neurons in each hidden layer is no less than the number of inputs plus three \citep{kidger2020universal}, we denote these as ``fixed-width" neural networks.

\citet{bruck1988power} showed, in a context different from ours, that polynomial-sized neural networks cannot solve \cNP-hard problems unless \cNP=co-\cNP\ and that polynomial-sized neural networks cannot solve the travelling salesperson problem $\epsilon$-approximately unless \cP=\cNP. \citet{helm2008} showed that A* with heuristic function values that underestimate the completely informed ones by at most a small constant must expand a number of states that scales exponentially with the instance size in several common planning problems.

\section{Feasibility of Approximating Completely Informed Heuristic Functions} \label{sec:theory}

The following corollary of the universal function approximation theorem shows that it is possible to use universal function approximators to approximate completely informed heuristic functions with high precision.

\begin{corollary} \label{lem:can_fit}
A universal function approximator is able to approximate the completely informed heuristic function with any desired precision for finite state spaces with a bounded informed heuristic range.
\end{corollary}

\begin{proof}
A polynomial (whose degree is the number of states minus one) can be used to represent the completely informed heuristic function. A universal function approximator can approximate this polynomial with arbitrary precision according to the universal function approximation theorem~\citep{hornik1989multilayer}.
%\SP{I don't agree with statement 1 of the proof. Please see original proof, a data of n dimensional can be fitted with n+1 degree polynomial. In my opinion its not very clear}
%\Taoan{The number of dimensions does not matter. We can map the state space $S$ to integers $\mathbb{N}$ and fit a  single variate polynomial (whose degree is the number of states minus one)  on the integers to the h values.}
%\SK{I don't know anything about the universal function approximation theorem. So, someone else needs to write this text to guarantee correctness.}
%Sumedh's post comment: maybe I read it wrong during transit. It makes sense to me now.
\end{proof}

% IS THIS CORRECT? ALSO, I CHANGED "IS THE NUMBER OF STATES PLUS ONE" TO "IS THE NUMBER OF STATES MINUS ONE" [Taoan: I think it is correct].

% Sven: It seems the informed heuristic function values of the states are trivially bounded if the number of states is finite? Also, I deleted that the state space needs to be discrete since finite state spaces have to be discrete.

% See Appendix~\ref{adx:can-fit} for the proof.

Consider the reduction of an \cNP-hard problem to a minimum-cost path problem. Each instance of the problem is reduced to a graph $G=(S,E)$ with positive edge costs, a start state in $S$, and a goal state in $S$ such that the cost of a minimum-cost solution of the instance is equal to the goal distance of the start state. The reduction needs to have the following properties. Property 1: The branching factor and the number of edges in each minimum-cost path from the start state to the goal state grows only polynomially in the size of the corresponding instance. Property 2: All edge costs are multiples of $\epsilon$ for a given constant $\epsilon > 0$.

For example, the standard reduction of the sliding tile puzzle (``Tile Puzzle") problem to the minimum-cost path problem satisfies Property 1 since the branching factor is bounded by four and the decision version of Tile-Puzzle is NP-complete \citet{RATNER1990111}, meaning that its solutions can be verified in time polynomial in the instance size. It also satisfies Property 2 for $\epsilon = 1$ since one minimizes the number of tile moves (meaning that each edge cost is one).

% Sven: I deleted tile-puzzles here since they have not been mentioned earlier. It is okay to include them here, but then they should be mentioned under related work in the main text.

In the following, we consider any NP-hard problem and its reduction that satisfy the above two properties. We say that an approximation $\hat{h}$ of the completely informed heuristic function $h^*$ has precision $p$ iff $\hat{h}(s) \in [h^*(s) - p, h^*(s) + p]$ for all states $s$. 

% Sven: I don't know why we need this. Isn't the A* example sufficient?
% The following theorem shows that finding an approximation of the completely informed heuristic function with precision higher than $\epsilon/2$ and calculating its function values in time polynomial in the instance sizes likely cannot be done both in polynomial time in the instance size.

% \begin{theorem}
% Consider any NP-hard problem and its reduction that satisfy Definition \ref{def:bnpc}. Then, calculating $\hat{h}(s)$ for the possible start states $s$ with precision higher than $\epsilon/2$ is NP-hard.
% \end{theorem}

% \begin{proof}
% If one has $\hat{h}(s)$ with the desired precision available, then the goal distance of the start state $s$ is known since only one multiple of $\epsilon$ is in the interval $(\hat{h}(s) - \epsilon/2, \hat{h}(s) + \epsilon/2)$. Thus, decision version of the NP-hard problem can be answered trivially since the minimum-cost solution of the instance is equal to the goal distance of the start state. 
% \end{proof}

Approximating an informed heuristic function with precision $p < \epsilon/2$ and precision $p=0$ are equivalent since $h^*(s)$ is the only multiple of $\epsilon$ that is in the interval $(\hat{h}(s) - \epsilon/2, \hat{h}(s) + \epsilon/2)$ according to Property 2.
%If both training a universal function approximator (that approximates the completely informed heuristic functions with precision $p \le \epsilon/2$) and calculating any single function value required  time only polynomial in the instance sizes, then one would be able to calculate $h^*(s)$ for any single state $s$ in time polynomial in the instance sizes since $h^*(s)$ is the only multiple of $\epsilon$ that is in the interval $(\hat{h}(s) - \epsilon/2, \hat{h}(s) + \epsilon/2)$ according to Property 2. 
As described earlier, the runtime of A* with the completely informed heuristic function (and appropriate tie-breaking) is polynomial in the branching factor and the number of edges in a minimum-cost path and thus polynomial in the instance sizes according to Property 1. Thus, if $\cP \ne \cNP$, then it is impossible to calculate each value of a high-precision approximation of the completely informed heuristic function in time polynomial in the instance sizes.

\section{Experiments}

For neural-networks, the runtime complexity for computing a value approximation (feedforward pass) is polynomial in the number of parameters.
As a result, we now study experimentally how the numbers of parameters (that is, weights and biases) of neural networks (as an informed heuristic approximator) increase with the instance sizes of \cNP-hard minimum-cost path problems and the precision of the approximations of the completely informed heuristic functions.

\subsection{Experimental Set-Up}

We use problem-specific encodings of three NP-hard minimum-cost path problems that satisfy the aformentioned reduction conditions, namely the pancake sorting ("Pancake"), travelling salesperson ("TSP"), and blocks world ("Blocks World") problems. The instance size is the number of pancakes, cities, and blocks, respectively. We could prune their state spaces, following \citep{valenzano2017analysis,fitzpatrick2021learning,slaney2001blocks}, but avoid so to show general trends and simplify the replication of our experiments by others. See Appendix~\ref{adx:domains} for more details on the problems. 

% Not sure about this sentence since a factorial growth actually seems to be pretty severe: Their numbers of states grow (relatively) slowly in their instance sizes, which allows us to obtain experimental results for many instance sizes. 

We use both fixed-depth and fixed-width neural networks as universal function approximators for the completely informed heuristic functions, following \citep{arfaee2010bootstrap,agostinelli2019deepcubea,agostinelli2021search}, in ways that satisfy the properties required by the universal function approximation theorem \citep{sonoda2017neural,lin2018resnet,kidger2020universal}. Both variants of neural networks use Rectified Linear Unit (ReLU) activation functions\footnote{ReLU activation functions have the properties required by the universal function approximation theorem~\citep{sonoda2017neural}.} \citep{nair2010rectified}. We also use residual connections\footnote{Residual networks are universal function approximators~\citep{lin2018resnet}.} \citep{he2016deep} and batch normalization~\citep{ioffe2015batch} to mitigate issues observed during the experiments, such as vanishing gradients. We do not use the convolutional neural networks in \citep{orseau2021policy} because their spatial locality assumptions do not hold in our state spaces.

Our fixed-width neural networks are similar to the ones in \citep{agostinelli2019deepcubea}, with two differences. First, we reduce the number of neurons per layer to allow for a more gradual increase in the number of parameters as the number of hidden layers increases. We set the number of neurons per hidden layer to the number of input dimensions plus three, following \citet{kidger2020universal}. Second, we have just one hidden layer (instead of two) before the residual blocks to allow for an odd number of hidden layers and, again, a more gradual increase in the number of parameters. 

We train the neural networks with the Adam optimizer \citep{kingma2014adam}, following \citep{agostinelli2019deepcubea,orseau2021policy}. It is unrealistic to train on the entire state space since the numbers of states increase exponentially in the instance sizes. Thus, we randomly select one million states and their completely informed heuristic function values for each instance size (with replacement) and use 80\% of them for the training set and the remaining 20\% for the test set. See Appendix~\ref{adx:domains} for details on generating the training and test sets.

Larger fixed-depth and fixed-width neural networks are able to achieve at least the precision of smaller fixed-depth and fixed-width neural networks, respectively, since the topologies of the former neural networks include the topologies of the latter ones. We thus use binary search on the number of neurons per hidden layer and the number of hidden layers, respectively, to determine the neural network with the smallest number of parameters that achieves the desired precision on the training set given by a problem-specific loss threshold.\footnote{We choose the loss thresholds so that they are sufficiently small to result in an increase in the number of parameters for small increasing instance sizes and sufficiently large to allow us to find neural networks that achieve the desired precision in a reasonable amount of time for large instance sizes.} The loss function is the mean squared error loss, following \citep{agostinelli2019deepcubea,agostinelli2021obtaining,agostinelli2021search,orseau2021policy}. Since the Adam optimizer is not guaranteed to achieve the highest possible precision, we repeat the binary search five times and return the neural network with the fewest parameters found.

The binary search runs the Adam optimizer until a desired precision or 300 epochs have been reached. In the former case, it remembers the current neural network as one that achieves the desired precision and continues the binary search to determine whether a smaller neural network is also able to achieve the desired precision. In the latter case, it continues the binary search to determine whether a larger neural network is able to achieve the desired precision.

\subsection{Experimental Results}

\begin{figure*}[!ht]
\begin{subfigure}{.32\textwidth}
  \centering
  % include first image
  \includegraphics[width=0.9\linewidth]{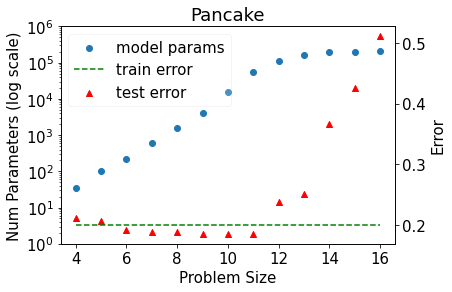}  
  %\caption{Put your sub-caption here}
  \label{fig:mse-pancake-0.2}
\end{subfigure}
\begin{subfigure}{.32\textwidth}
  \centering
  % include second image
  \includegraphics[width=0.9\linewidth]{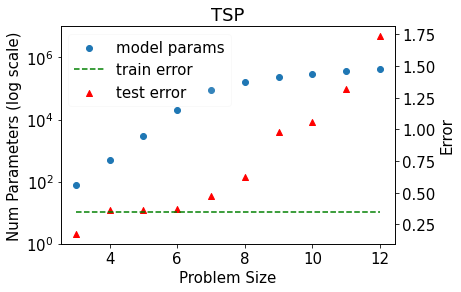}  
  %\caption{Put your sub-caption here}
  \label{fig:mse-tsp-35}
\end{subfigure}
\begin{subfigure}{.32\textwidth}
  \centering
  % include first image
  \includegraphics[width=0.9\linewidth]{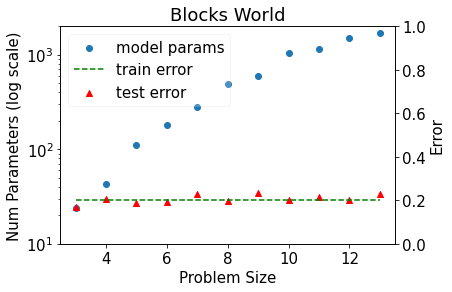}  
  %\caption{Put your sub-caption here}
  \label{fig:mse-blocks-0.1}
\end{subfigure}
\caption{The minimum numbers of parameters (in log scale) required for fixed-depth neural networks to approximate the training sets with given loss thresholds (left) and the resulting losses on the test sets (right) as functions of the instance sizes. The loss threshold is 0.2, 0.35, and 0.2 for Pancake, TSP, and Blocks World, respectively.}
\label{fig:results-param-log-fixed-depth}
\end{figure*}
\begin{figure*}[!ht]
\begin{subfigure}{.32\textwidth}
  \centering
  % include first image
  \includegraphics[width=0.83\linewidth]{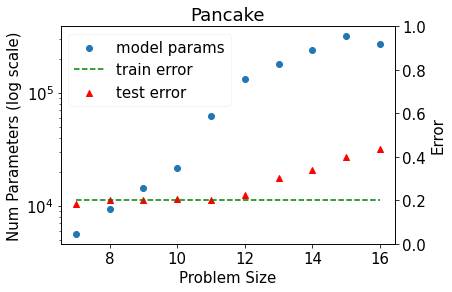}  
  %\caption{Put your sub-caption here}
  \label{fig:mse-pancake-fw-0.1}
\end{subfigure}
\begin{subfigure}{.32\textwidth}
  \centering
  % include second image
  \includegraphics[width=0.93\linewidth]{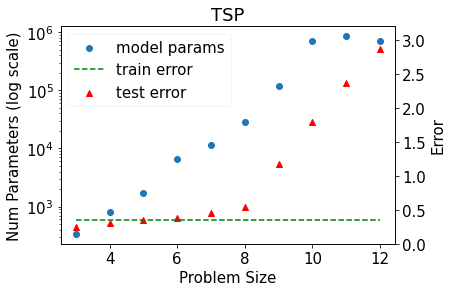}  
  %\caption{Put your sub-caption here}
  \label{fig:mse-tsp-35-2}
\end{subfigure}
\begin{subfigure}{.32\textwidth}
  \centering
  % include first image
  \includegraphics[width=0.93\linewidth]{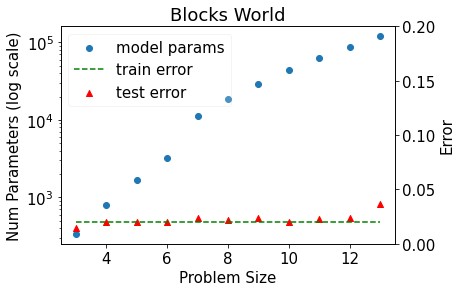}  
  %\caption{Put your sub-caption here}
  \label{fig:mse-blocks--2}
\end{subfigure}
\caption{The minimum numbers of parameters (in log scale) required for fixed-width neural networks to approximate the training sets with given loss thresholds (left) and the resulting losses on the test sets (right) as functions of the instance sizes. The loss threshold is 0.1, 0.35, and 0.02 for Pancake, TSP, and Blocks World, respectively.}
\label{fig:results-param-log-fixed-width}
\end{figure*}

%IN ALL PLOTS, USE "MODEL PARAMETERS" INSTEAD OF "MODEL PARAMS", "LOSS ON THE TRAINING SET" INSTEAD OF "TRAIN ERROR", "LOSS ON THE TEST SET" INSTEAD OF "TEST ERROR", "INSTANCE SIZE" INSTEAD OF "PROBLEM SIZE".

\paragraph{Experiment 1} Figure \ref{fig:results-param-log-fixed-depth} shows the smallest numbers of parameters required for fixed-depth neural networks to approximate the training sets with given loss thresholds (in log scale) and the resulting losses on the test sets as functions of the instance sizes. 

For Pancake and TSP, we first see linear trends for the number of parameters in log scale as their instance sizes increase (Phase 1), which suggests that the numbers of parameters increase exponentially in the instance sizes. We then see stagnations around Pancake instances of size 12 and TSP instances of size 7 (Phase 2). The losses on the test sets start to increase roughly at the point of stagnation. One explanation for the increasing losses on the test sets is that the training and test sets contain fewer and fewer common states as the instance sizes increase since the numbers of states increase with the instance sizes, and it thus becomes more and more unlikely that the same states will be part of both the training and test sets. Thus, the ability to generalize well to states not in the training set becomes more and more important, but the increasing losses on the test sets show that generalization is poor. The poor generalization could be due to overfitting the training sets, which we did not prevent in our experiments (other than by stopping training early) since we need to approximate the completely informed heuristic functions closely for all states, including those in the training sets. For Blocks World, we see only Phase 1 and thus cannot be sure whether it is followed by Phase 2 (overfitting). We tried to increase its instance sizes to ensure that it is not a counter example to our claims, but the generation of the training is not feasible given our available computing resources.

Figure \ref{fig:results-param-log-fixed-width} shows similar patterns for fixed-width neural networks although the graphs are noisier than those for fixed-depth neural networks since adding another hidden layer with $n$ neurons increases the number of parameters by $n^2+n$ and thus by more than for fixed-depth neural networks, resulting in less smooth graphs. 

Overall, our results suggest that completely informed heuristic functions for our \cNP-hard minimum-cost path problems might not have the necessary structure to achieve sufficiently compressed representations of the completely informed heuristic functions needed to scale to large instance sizes. In our experiments, we keep the numbers of states in the training sets constant, which appears to make the sizes of the neural networks stagnate as the instance sizes increase but also prevents them from approximating the completely informed heuristic functions with high precision.

% \subsubsection{Invariance of Experimental Results to the Loss Threshold}

\begin{figure*}[!ht]
\begin{subfigure}{.329\textwidth}
    \centering
    \includegraphics[width=0.9\linewidth]{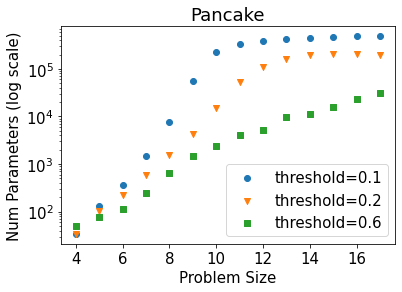}
\end{subfigure}
\begin{subfigure}{.329\textwidth}
    \centering
    \includegraphics[width=0.9\linewidth]{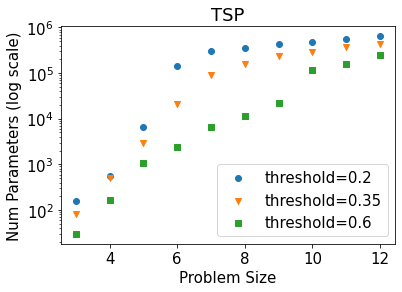}
\end{subfigure}
\begin{subfigure}{.329\textwidth}
    \centering
    \includegraphics[width=0.9\linewidth]{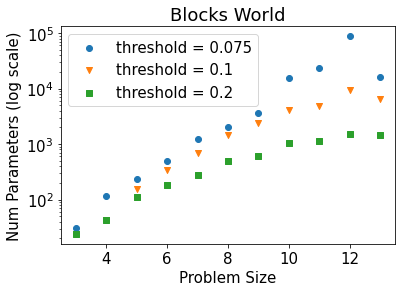}
\end{subfigure}
\caption{The minimum numbers of parameters (in log scale) required for fixed-depth neural networks to approximate the training sets  with three loss thresholds each as functions of the instance sizes.}
\label{fig:results-threshold}
\end{figure*}

% \SP{Is this caption sufficient?}

\paragraph{Experiment 2} So far, the loss thresholds were hand-selected for each problem. We now report experimental results for two additional loss thresholds per problem. 

Figure \ref{fig:results-threshold} shows that larger loss thresholds for fixed-depth neural networks result in smaller numbers of parameters since it is easier to approximate functions with larger loss thresholds. However, our experimental results for the numbers of parameters are similar for all loss thresholds. Our conclusions of Experiment 1 thus continue to apply, namely that there are linear trends for the numbers of parameters in log scale (possibly followed by stagnations) as the instance sizes increase.

%IF THERE IS A CHANGE IN OVERFITTING, IT IS CERTAINLY NOT SYSTEMATIC. FOR PANCAKE, IT SEEMS IT OCCURS LATER FOR LARGER LOSS THRESHOLDS. FOR BLOCKS WORLD, IT SEEMS IT OCCURS EARLIER. BUT DOES IT REALLY OCCUR? WHY DON'T YOU SHOW THE TEST ERROR? ALSO, WHAT IS GOING ON FOR HIGH INSTANCE SIZES FOR BLOCKS WORLD, WHERE THE CURVES START TO GO DOWN? I DON'T SEE YOUR STATEMENT THAT OVERFITTING OCCURS WHEN THE NUMBER OF PARAMETERS REACHES ABOUT $5x10^5$ FOR BLOCKS WORLD WITH LOSS THRESHOLD 0.2.

% \subsubsection{Invariance of Experimental Results to the Loss Function}

\begin{figure*}[htpb]
\begin{subfigure}{0.32\textwidth}
  \centering
  % include first image
  \includegraphics[width=0.9\linewidth]{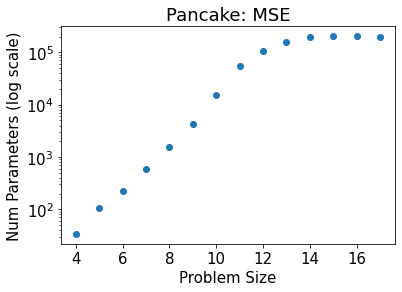}  
  %\caption{Put your sub-caption here}
\end{subfigure}
\begin{subfigure}{0.32\textwidth}
  \centering
  % include first image
  \includegraphics[width=0.9\linewidth]{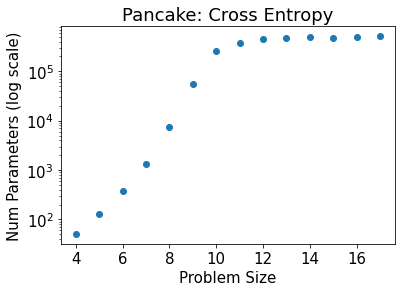}  
  %\caption{Put your sub-caption here}
\end{subfigure}
\begin{subfigure}{0.32\textwidth}
  \centering
  % include first image
  \includegraphics[width=0.9\linewidth]{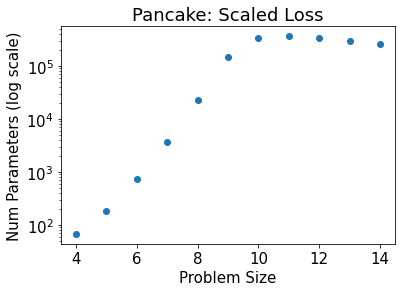}  
  %\caption{Put your sub-caption here}
\end{subfigure}
\caption{The minimum numbers of parameters (in log scale) required for fixed-depth neural networks to approximate the training sets for Pancake with given loss thresholds as functions of the instance sizes. The loss threshold of the MSE, cross entropy, and scaled loss is 0.2, 90\%, and 0.001, respectively.}
\label{fig:fig-scaled}
\end{figure*}

\paragraph{Experiment 3} So far, we treated the approximation of the training sets as a regression problem and used mean squared error as the loss function.  We now report experimental results for two additional settings, namely 1) treating the problem as a classification problem and using the (categorical) cross-entropy loss as loss function, following \citep{ferber2020neural}, and 2) using the ``scaled" loss as loss function, defined as 
$\frac{1}{N}\sum_{i=1}^{N} \left(\frac{\hat{h}(s_i)}{h^*(s_i)}-1\right)^2$,
where $N$ is the size of the training set and $s_i$ is the state of its $i$th element. The motivation behind the scaled loss is that the suboptimality factor of A* is governed by the relative error $\hat{h}(s_i)/h^*(s_i)-1$ rather than the absolute error $\hat{h}(s_i)-h^*(s_i)$ if $\hat{h}(s_i)>h^*(s_i)$ \citep{ebendt2009weighted}.

Figure~\ref{fig:fig-scaled} shows that our experimental results for the numbers of parameters of fixed-depth neural networks are similar for all loss functions. Our conclusions of Experiment 1 thus continue to apply, namely that there are linear trends for the numbers of parameters in log scale (possibly followed by stagnations) as the instance sizes increase.

\section{Conclusion}

We supported our claim that approaches that use universal function approximators to approximate completely informed heuristic functions with high precision to speed up A* searches for solving NP-hard minimum-cost path problems do not scale to large instance sizes. The research community might therefore want to investigate other ways of integrating heuristic search with machine learning.

% Use \bibliography{yourbibfile} instead or the References section will not appear in your paper

\clearpage

\bibliographystyle{plainnat}

\bibliography{aaai22}

\clearpage

%%%%%%%%%%%%%%%%%%%%%%%%%%%%%%%%%%%%%%%%%%%%%%%%%%%%%%%%%%%%
\appendix
\section{Summary of Previous Results}
\label{sec:related-summary}
\begin{table*}[!htpb]
\centering
\small
\begin{tabular}{lrlc@{}rc@{}rc@{}r}
\hline
& & & & & \multicolumn{4}{c}{Results of the Search} \\
(Inst. Size) Problem & \multicolumn{1}{l}{\# States} & Approach & \multicolumn{2}{l}{\# Parameters} & \multicolumn{2}{l}{Path Length} & \multicolumn{2}{l}{\# Expanded States} \\ \hline
\multirow{1}{*}{48 Tile Puzzle} & \multirow{1}{*}{$3.00 \times 10^{62}$} & DeepCubeA & & $3.00 \times 10^7$ & & $253.4$ & & *$5.73 \times 10^{6}$ \\ \hline
\multirow{3}{*}{24 Tile Puzzle} & \multirow{3}{*}{$7.70 \times 10^{24}$} & DeepCubeA & $\leq$ & $2.10 \times 10^7$ & $\geq$ & $89.5$ & $\leq$ & *$2.01 \times 10^{6}$ \\
& & PHS* & & $1.05 \times 10^6$ & $\leq$ & $224.0$ & $\geq$ & $2.87 \times 10^{3}$ \\
& & PHS$_h$ & & $1.05 \times 10^6$ & & $119.5$ & & $5.86 \times 10^{4}$ \\
& & BLH & $\geq$ & *$3.00 \times 10^4$ & & - & & $5.22 \times 10^{6}$ \\ \hline
\multirow{2}{*}{15 Tile Puzzle} & \multirow{2}{*}{$1.00 \times 10^{13}$} &  DeepCubeA & $\leq$ & $1.82 \times 10^7$ & & $52.0$ & $\leq$ & $*1.28 \times 10^{6}$ \\
& & BLH & $\geq$ & *$3.00 \times 10^4$ & & - & $\geq$ & $1.01 \times 10^{4}$ \\ \hline
\multirow{3}{*}{Sokoban} & \multirow{3}{*}{*$1.53 \times 10^{15}$} & DeepCubeA & $\leq$ & $1.50 \times 10^7$ & $\geq$ & $32.9$ & $\geq$ & $1.05 \times 10^{3}$  \\
& &  PHS* & $\geq$ & $3.71 \times 10^6$ & & $37.6$ & & $1.52 \times 10^{3}$ \\ 
& &  PHS$_h$ & $\geq$ & $3.71 \times 10^6$ & $\leq$ & 39.1 & $\leq$ & $2.13 \times 10^{3}$ \\ \hline
\end{tabular}

\caption{Comparison of previous approaches: DeepCubeA~\protect\citep{agostinelli2019deepcubea}, Policy Guided Heuristic Search (PHS*, PHS$_h$)~\protect\citep{orseau2021policy}, and Bootstrap Learning Heuristic (BLH) ~\protect\citep{arfaee2010bootstrap}. "-" denotes an unreported value in the original paper. "*" in front of a value denotes that it is an approximation based on additional assumptions. "$\leq$" in front of a value denotes that it is the worst observed value. "$\geq$" in front of a value denotes that it is the best observed value.
}
\label{tab:comparison-previous}
\end{table*}

Table~\ref{tab:comparison-previous} shows a summary of some existing experimental results from the papers of previous approaches. We estimate the number of states of Sokoban on a $10 \times 10$ grid with 4 boxes as $100 \times {100 \choose 4} \times {100 \choose 4}$, which is the product of the numbers of possible player locations, possible start locations of the boxes, and possible goal locations of the boxes. We estimate the numbers of parameters of BLH by assuming neural networks with 2 hidden layers and 1,000 neurons per hidden layer. In general, we ignore all parameters of the batch normalization layers. Changing numbers of parameters for an approach result from changing numbers of inputs due to the changing instance sizes rather than any changes to the layouts of the neural networks. An exception is DeepCubeA, which uses 6 residual blocks \citep{he2016deep} for Tile Puzzle and 4 residual blocks for Sokoban. We estimate the numbers of expanded states of DeepCubeA by dividing the reported numbers of generated states by the branching factors.

Our research was motivated by the observation that, for Tile Puzzle, DeepCubeA produces neural networks with larger numbers of parameters for larger instance sizes.

% Our research is motivated by the observation that, for Tile Puzzle, the same approach results in larger numbers of parameters (as is the case for DeepCubeA) or more expanded states (as is the case for BLH) for larger instance sizes.
% \SK{So, I wrote something in the paragraph above trying to follow Sumedh's argument. My text makes sense to me for the "larger numbers of parameters," but not for the "more expanded states". I would expect that, in larger state spaces, the resulting paths will be longer and thus more states get expanded. This does not necessarily need to have anything to do with the heuristics being approximated less closely!}

% \SP{The results suggest that there might be a trend where scaling the problem size necessitates a larger approximator (w.r.t the number of parameters). That is, if the approximator's size is not increased, then we observe reduced accuracy (`Solution Quality') and/or increased computational complexity (`Expanded States').} \SK{I don't understand the suggestion. What I understand does not fit the experimental results in the table. Let's talk on Zoom on Monday.}

% I DON'T KNOW WHAT THE CONCLUSIONS FROM THE PREVIOUS RESULTS SHOULD BE. THEY COULD JUST BE DELETED. IF THEY REMAIN, THEN THE RUBIK'S CUBE SHOULD BE ADDED SINCE WE MENTION IT TWICE IN THE RELATED WORK SECTION. WE MENTION THE WORK BY MCALEER IN THE RELATED WORK SECTION IN THE CONTEXT OF THE RUBIK'S CUBE. SO, SHOULDN'T THAT WORK AND DEEPCUBEA SHOW UP IN THE TABLE FOR THE RUBIK'S CUBE?

\begin{figure*}[!htpb]
\begin{subfigure}{.32\textwidth}
    \centering
    \includegraphics[width=0.94\linewidth]{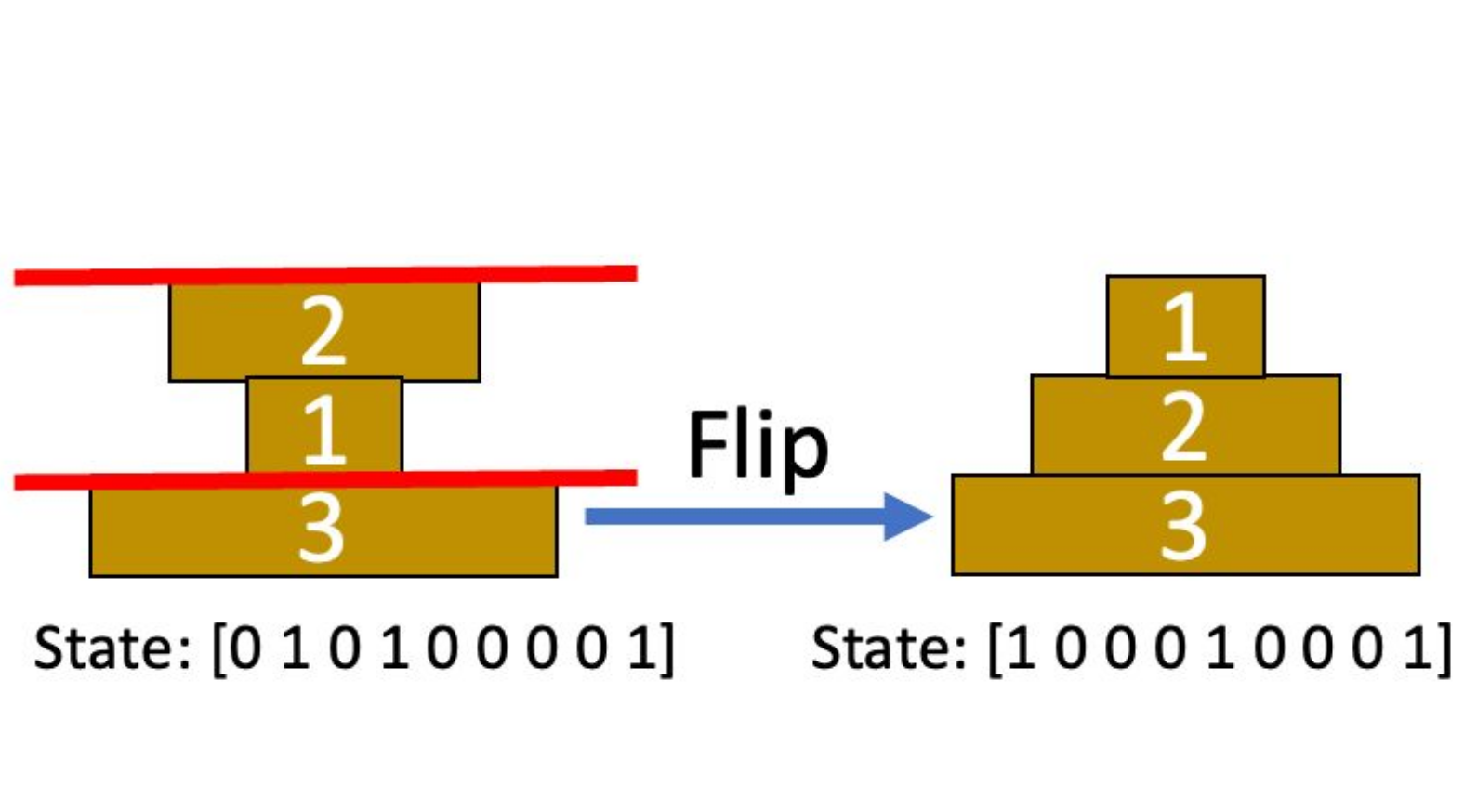}
    \label{fig:domain-pancake}
    \subcaption[]{Pancake}
\end{subfigure}
\begin{subfigure}{.329\textwidth}
    \centering
    \includegraphics[width=.55\linewidth]{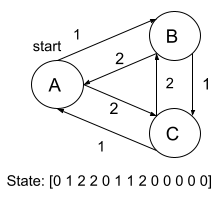}
    \label{fig:domain-tsp}
    \subcaption[]{TSP}
\end{subfigure}
\begin{subfigure}{.329\textwidth}
    \centering
    \includegraphics[width=.7\linewidth]{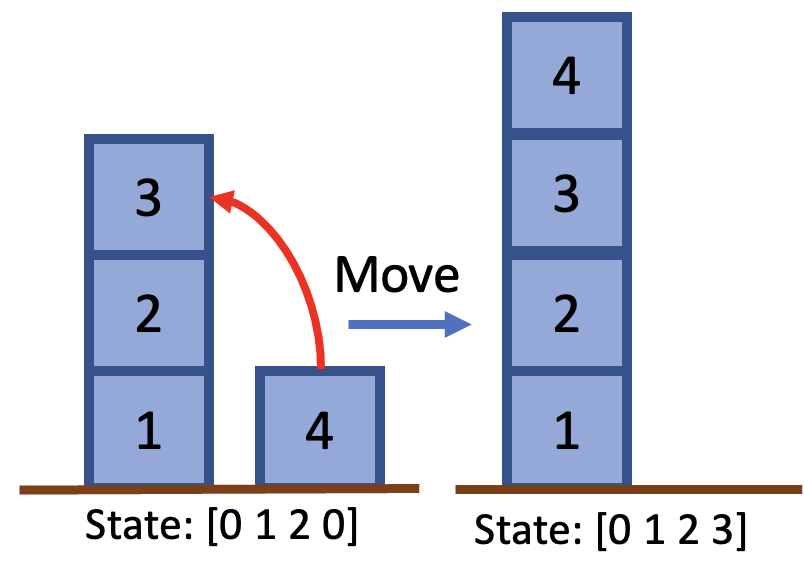}
    \label{fig:domain-bw}
    \subcaption[]{Blocks World}
\end{subfigure}
\caption{States and their encodings for the three NP-hard minimum-cost path problems.}
\label{fig:domains}
\end{figure*}

\section{Description of the NP-Hard Minimum-Cost Path Problems} \label{adx:domains}

\paragraph{Pancake} 
(1) \textit{Description:} The pancake sorting problem ("Pancake") is \cNP-hard~\citep{bulteau2015pancake}. The task is to sort pancakes of different sizes, that are stacked on top of each other, with the minimum number of moves, each of which consists of inserting a spatula at any position in the stack and flipping (inverting) all pancakes above it. 
(2) \textit{Dataset Generation:} We generate the training sets via a random walk from the goal states, following~\citep{agostinelli2019deepcubea}. We calculate the completely informed heuristic function values of the encountered states via an A* search with the consistent gap heuristic function~\citep{helmert2010landmark}.
(3) \textit{Encoding:} We encode a state via a one-hot encoding of the location of each pancake in the stack, as shown in Figure~\ref{fig:domains}(a).
(4) \textit{Properties 1 and 2:} Pancake satisfies the properties for $\epsilon = 1$ since the branching factor is the number of pancakes minus one, the number of edges in a minimum-cost path is at most twice the number of pancakes (since one can solve any Pancake instance by repeatedly bringing the largest pancake not in its correct position in the goal configuration to the top of the stack with one flip and then to its correct position in the goal configuration with another flip), and one minimizes the number of flips (meaning that each edge cost is one).

\paragraph{TSP}
(1) \textit{Description:} The asymmetric travelling salesperson problem ("TSP") is  NP-hard~\citep{cormen2009introduction}. The task is to find a minimum-cost tour that visits every city.
(2) \textit{Dataset Generation:} We generate the training sets by considering (only) the start states of the searches (where the salesperson is still in the start city) of different TSP instances of the same instance size.\footnote{It is future work to repeat our experiments with training and test sets that contain not only the start states of the searches.} We generate complete weighted directed graphs with the given number of cities and edge costs that are uniformly sampled from 0.1, 0.2 \ldots 50.0. We calculate the completely informed heuristic function values of the start states via the Held–Karp algorithm~\citep{held1962dynamic}. (3) \textit{Encoding:} We encode a state as a vector that encodes, for each edge, its cost and, for each city, whether it has already been visited or is currently visited, as shown in Figure~\ref{fig:domains}(b). The first city listed is the start city. 
(4) \textit{Properties 1 and 2:} Our version of TSP satisfies the properties for $\epsilon = 0.1$ since the branching factor is the number of cities minus one, the number of edges in a minimum-cost path equals the number of cities, and each edge cost is a multiple of 0.1.

% \SK{What is learned for TSP does not support A* searches and thus does not fit the premise of the paper since we learn an approximations of the completely informed heuristic function values only for the start states of the searches but no other states!}

\paragraph{Blocks World} 
(1) \textit{Description:} The Blocks World ("BW") problem is NP-hard ~\citep{gupta1992complexity}, and we suspect that it remains NP-hard with the fixed goal configuration that we use. Several toy blocks, numbered $1 \ldots n$, form stacks on a table top. The task is to transform a given start configuration of blocks into a given goal configuration with the minimum number of moves, each of which moves a block from the top of a stack to the top of another stack or the table. 
(2) \textit{Dataset Generation}: We use the goal configurations where all blocks are stacked in order of their numbers in one stack, with block 1 at the bottom. We generate the training sets by randomly generating start configurations. We calculate their completely informed heuristic function values via an A* search with the consistent heuristic function given by the numbers of blocks that are directly on top of blocks or the table top that they are not directly on top of in the goal configuration. 
(3) \textit{Encoding}: We encode a state as a vector that encodes, for each block, which block it is directly on top of (with 0 denoting the table top), following \citep{slaney2001blocks}, as shown in Figure~\ref{fig:domains}(c).
(4) \textit{Properties 1 and 2:} Blocks World satisfies the properties for $\epsilon = 1$ since the branching factor grows quadratically in the number of blocks, the number of edges in a minimum-cost path is at most twice the number of blocks (since one can solve any Blocks World instance by first moving all blocks directly onto the table top and then to their correct positions in the goal configuration), and one minimizes the number of block moves (meaning that each edge cost is one).

% \SK{To model the Blocks World problem as a search problem, we need to assume that the goal is fixed - in other words, the goal state is not part of the definition of an instance. For example, in our experiments, the goal configurations consists of one stack that contains all blocks in order of their numbers. The cited paper does not seem to state that this problem is NP-hard and/or its decision version is NP-complete, although I suspect that this is the case. But we cannot make these claims without evidence, especially since there are goal configurations where this is clearly not the case (for example, the goal configuration where all blocks are directly on the table top).}

\section*{Acknowledgments}

The research at the University of Southern California was supported by the National Science Foundation (NSF) under grant numbers 1409987, 1724392, 1817189, 1837779, 1935712, 2121028, and 2112533. The views and conclusions contained in this document are those of the authors and should not be interpreted as representing the official policies, either expressed or implied, of NSF or the U.S. government.

\end{document}